\documentclass[a4paper,11pt]{article}
\usepackage{CLARIN2015}

\usepackage[T1]{fontenc} 
\usepackage{times}
\usepackage{url}
\usepackage{latexsym}
\usepackage[english]{babel}
\usepackage{graphicx,hyperref}

\usepackage{covington} 

\title{Polish Read Speech Corpus for Speech Tools and Services}

\author{Danijel Kor\v{z}inek \and Krzysztof Marasek \and \L{}ukasz Brocki \and Krzysztof Wo\l{}k \\
 Polish-Japanese Academy of Information Technology, \\
 Warsaw, Poland \\
 {\tt (danijel,kmarasek,lucas,kwolk)@pja.edu.pl} \\
}

\begin{document}
\maketitle
\begin{abstract}

This paper describes the speech processing activities conducted at the Polish consortium of the CLARIN project. The purpose of this segment of the project was to develop specific tools that would allow for automatic and semi-automatic processing of large quantities of acoustic speech data. The tools include the following: grapheme-to-phoneme conversion, speech-to-text alignment, voice activity detection, speaker diarization, keyword spotting and automatic speech transcription.  Furthermore, in order to develop these tools, a large high-quality studio speech corpus was recorded and released under an open license, to encourage development in the area of Polish speech research. Another purpose of the corpus was to serve as a reference for studies in phonetics and pronunciation. All the tools and resources were released on the the Polish CLARIN website. This paper discusses the current status and future plans for the project.

\end{abstract}

\blfootnote{This work is licenced under a Creative Commons Attribution 4.0 International Licence. Licence details: \url{http://creativecommons.org/licenses/by/4.0/}}

\section{Introduction} \label{intro}

Much of the data used in Humanities and Social Sciences (HSS) research is stored in the form of audio recordings. Examples of this include radio and television programmes, interviews, public speeches (e.g. parliament, public events), lectures, movies, read literary works and other recordings of speech. This data contains valuable information from many aspects of HSS research. This encompasses both the linguistic (with the emphasis on vocabulary and pronunciation) and sociological (emphasis on speaker) points of view. During our project, we have met many scientists that have shown interest in processing either already available data and corpora, or would like to process recordings (e.g. interviews) they intended to make in the future.

The main issue with processing acoustic data is that it is more expensive and time consuming than, for example, traditional, textual data. It demands both the know-how and lots of effort to achieve comparable results. That is why it is often overlooked by researchers who either do not have the time or the funding to deal with such issues. Our primary goal was to create free and accessible solutions for researchers from the HSS community.

Similar efforts in other Clarin consortia already exist, like WebMAUS \cite{kisler2016} speech segmentation services at LMU, AVATech \cite{lenkiewicz2012avatech} by Max Planck Institute and Fraunhofer Institute which provide video and audio processing services including speech segmentation, VAD and speaker diarization, and TTNWW \cite{clarinnl-ttnww} which includes speech transcription services for Dutch. It is worth noting that many of them (although not all) are language dependent, requiring a re-implementation of these services in individual countries.

This paper will first describe the tools and corpora created during the project. Next, it will describe a few of the existing and planned applications of the tools and services. Finally, it will describe the plans of development for the upcoming years.

\section{Speech Tools}

One of the earliest decision during the project was to release all the tools in the form of web services, rather than downloadable applications. There are many advantages to this: ease of use (no installation required), better support, stable environment and performance. However, a few disadvantages as well: more effort required from the consortium, increased response time if many people use the platform, issues with releasing sensitive data. Most of these have been addressed individually and by releasing the source code of the tools for ambitious individuals.

The main website located at \url{http://mowa.clarin-pl.eu} was divided into three sections: speech corpora downloads, grapheme-to-phoneme (G2P) conversion and the rest of the speech processing tools. The reason for removing the G2P from the rest of the tools is because it uses a different set of modalities (i.e. text-to-text) from the rest of the tools (audio and optionally text into text).

\begin{figure}
    \centering
    \includegraphics[width=.9\linewidth]{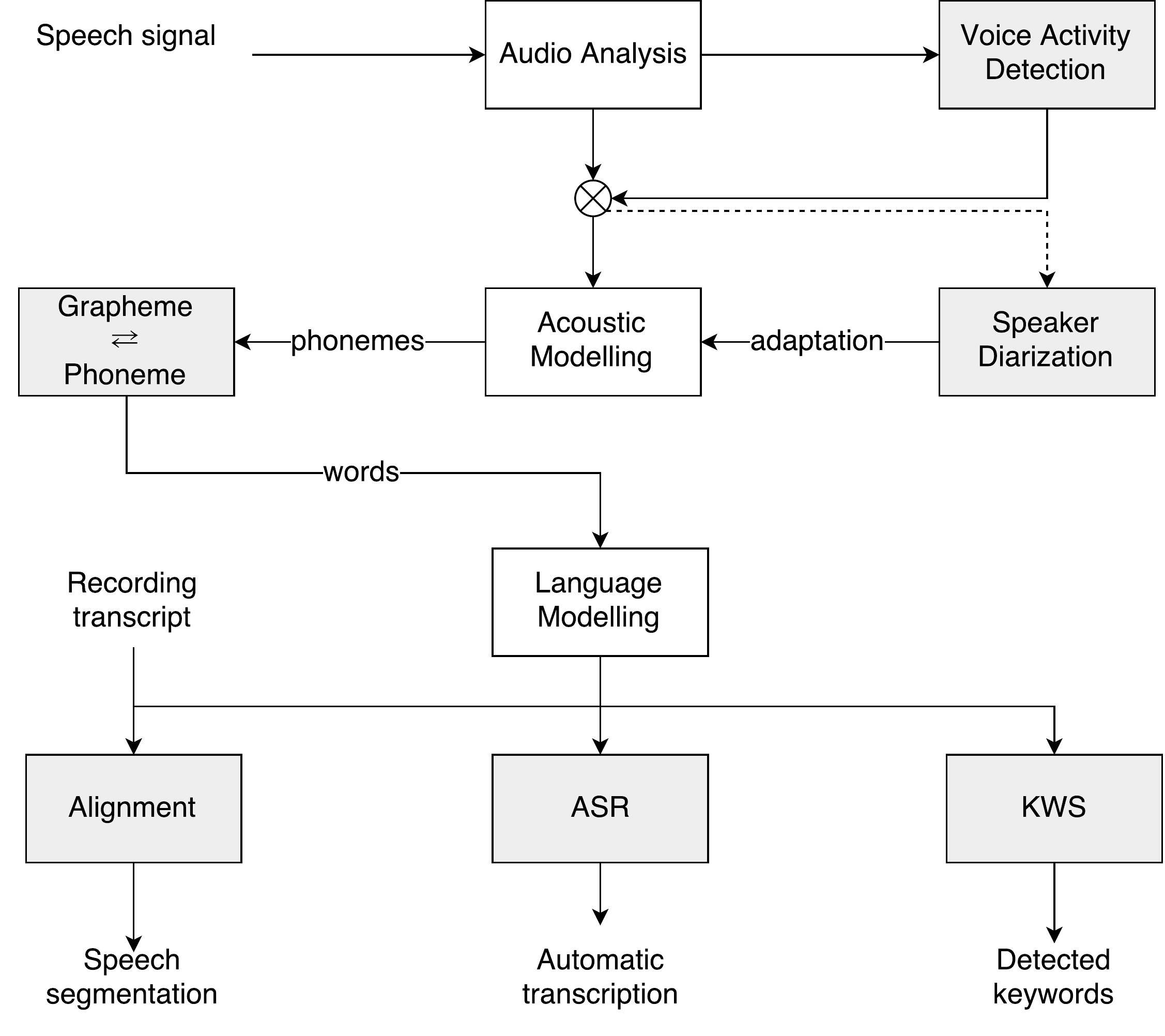}
    \caption{The pipeline of several speech processing mechanisms. The input signal on top is processed to produce the information on the bottom. The grey blocks were exposed as services in the CLARIN-PL infrastructure.}
    \label{fig:ASRtech}
\end{figure}

The tools were selected from the basic pipeline used in speech processing and speech recognition. Rather than hiding them, it was decided to expose the intermediary steps in the pipeline as standalone services. Figure \ref{fig:ASRtech} shows a typical speech processing pipeline used to produce the information shown at the bottom of that graph. The grey-filled blocks are the services that were exposed as standalone services. The remaining three blocks were omitted either because they were already available in other forms (language modeling) or because they were deemed not too useful for HSS research (audio analysis, acoustic modeling).

\subsection{Grapheme-to-phoneme conversion}

This tool allows converting any text written in the orthographic (i.e. written) form into its phonetic (i.e. spoken) form. It is one of the primary steps in any process that involves speech data. but may also serve as a tool outside of the acoustic speech processing context.

\begin{figure}
    \centering
    \begin{verbatim}
    f S tS e b Z e S I ni e x S on S tS b Z m i ft S tsi i ni e 
    i S tS e b Z e S I n s t e g o s w I ni e v u w g o p I t a 
    p a ni e x S on S tS u p o ts u S p a n t a g b Z en tS I 
    v g on S tS u
    \end{verbatim}
    \caption{An example of a transcription of the poem ``Chrz\k{a}szcz'' by Jan Brzechwa, as produced by the grapheme-to-phoneme tool. Uses a slightly modified version of the Polish SAMPA phonetic alphabet.}
    \label{fig:g2p}
\end{figure}

The tool is created using a rule-based system. It accepts any form of text, although it does not perform text normalization (it does not expand numbers, dates or abbreviations automatically). The system is completely rule based and contains a list of exceptions for names, foreign and other atypical words. A statistical system, based on the Sequitur \cite{bisani2008joint} tool, is also available but due to available data, it does not outperform the rule based system in any way. The tool can generate both word lists (with multiple pronunciations) and a canonical transcription of the text.

There are several enhancements planned for the future of this tool. The foremost includes the normalization, mentioned above. This problem is particularly difficult for inflected languages (e.g. Slavic languages). This work is already in progress as of writing this paper \cite{brocki2012multiple}. Other improvements would include adding different forms of phonetic alphabets (while retaining the same rule-set) and possibly adding other levels of annotation (e.g. accents and syllabification). Some uses may also benefit from using a graph-based representation, but these (and other) improvements will be added pending further interest from the community.

\subsection{Speech-to-text alignment}

Speech alignment is one of the most useful tools available. It is used to align a sequence of words to the provided audio recording of speech. This can be understood simply as automatically generating a set of time-codes, when both the audio and its transcription are known. It is a very useful tool because it can be used to easily look up specific events in large sets of recordings. It also makes possible to compute statistics related to the duration and other characteristics of individual speech events.

\begin{figure}
    \centering
    \includegraphics[width=.9\linewidth]{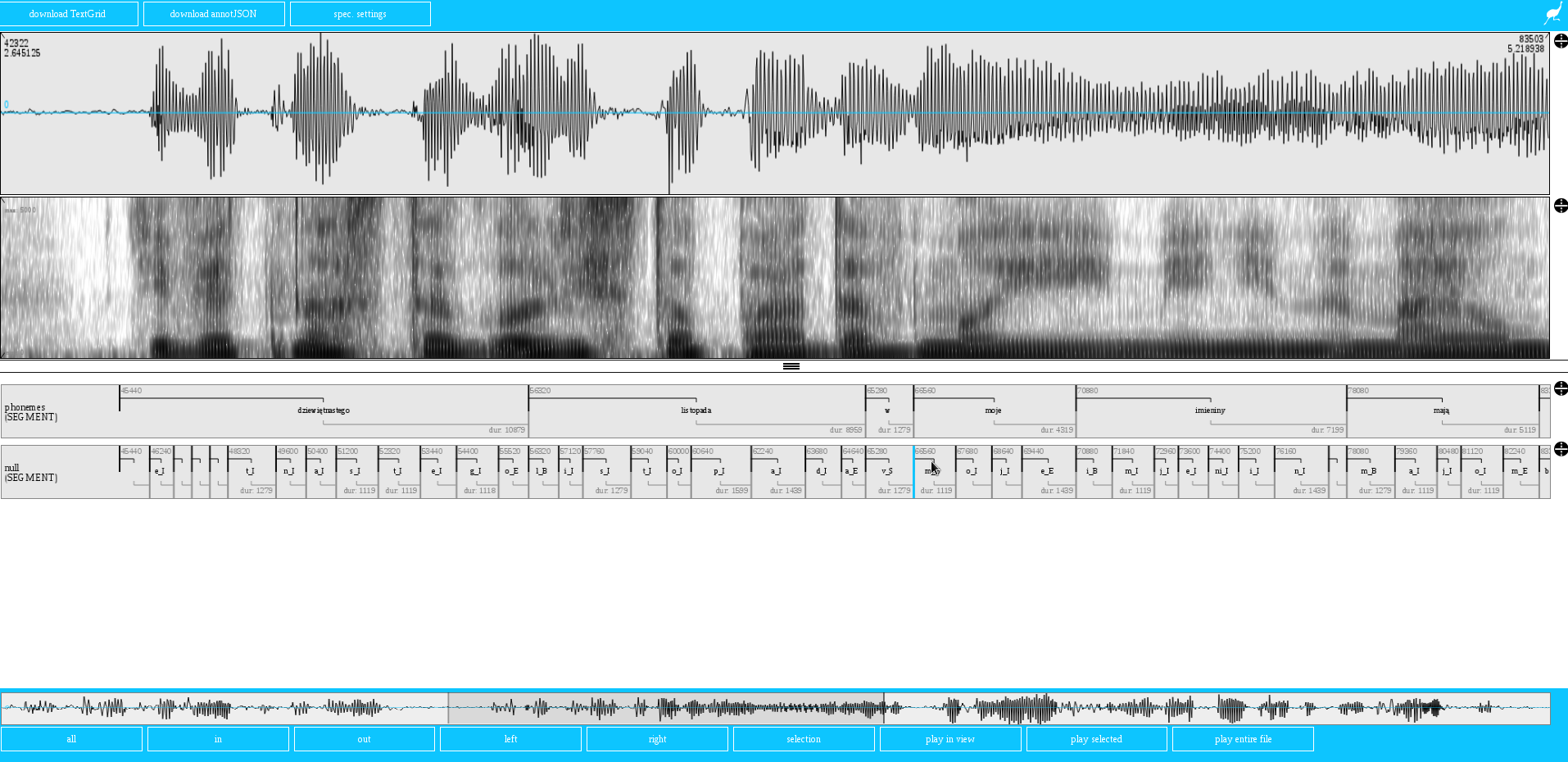}
    \caption{An example of a segmentation generated by the speech-to-text alignment service, displayed in the EMU-webApp interface.}
    \label{fig:emu}
\end{figure}

The tool was created, based on the SailAlign \cite{katsamanis2011sailalign} concept, in order to work efficiently with long audio files. The engine is constructed around the Kaldi toolkit \cite{Povey_ASRU2011}, just like most of the tools in this paper, but the main work-flow is managed using a set of libraries written in Java. The alignment is produced both on the level of words and phonemes. The tool currently only generates outputs in the form of a Praat TextGrid file \cite{boersma2002praat}, but others could be added in the future. The service also generates a link to the new EMU-webApp website \cite{winkelmann2014introducing}, which allows viewing the result of the segmentation directly in the browser (see figure \ref{fig:emu}).

For the future, several improvements are planned. A better acoustic model, possibly based on ANNs is going to be implemented. Adaptation of both the acoustic and the language model could also be beneficial to the overall process, especially when it comes to noisy data. The tool works fine for clean and predictable data, but it still produces errors or fails entirely for very noisy or otherwise low SNR signals. 

Finally, a UI enhancement could be created to allow the user to manually invoke the re-alignment of a certain portion of the output, while also allowing to fix some of the information (like the orthographic or the phonetic transcription). This could turn the already very useful, but sometimes imperfect fully automatic tool, into a perfect semi-automatic tool.

\subsection{Voice activity detection}

Voice activity detection (VAD) is frequently found as a pre-processing step of many speech processing tools. Its purpose is to isolate portions of audio that contain speech from those that contain other types of acoustic events, like silence, noise or music. Apart from the aforementioned use as a pre-processing step, it can also be useful as an indexing tool for large quantities of audio. This tool is completely language and domain independent, although it may fail with very noisy data.

This tool was constructed using an Artificial Neural Network based classifier that performs VAD in an online manner. The non-speech data is further analyzed using an SVM classifier to try and classify types of noise. This last step was not developed very thoroughly and performs rather poorly, depending on the data, but given a proper use-case and training data, it could be modified to work better.

The VAD component was used extensively during previous projects and was already known to perform reasonably well for real-world data. A simple experiment confirmed a fairly high level of recall ($\sim$99\%) with a not so good precision ($\sim$58\%), which was a conscious decision (preferring not to lose any speech, while sometimes accepting non-speech falsely). The subsequent tools (ASR modules) can deal with a small amount of noise in the data, but would suffer greatly if any speech was omitted.

\subsection{Speaker diarization}

This tool is used to segment a large audio file into portions spoken by individual speakers. There are several types of speaker related segmentation strategies that can be performed: speaker change detection recognizes only the segments where different speakers are talking, speaker diarization additionally annotates which segments belong to the same speaker and speaker identification recognizes exactly who the speaker talking in each segment is (e.g. by name). Our tool only does the second algorithm. 

It is mostly useful for adaptation of various tools and models to individual speakers but some researchers have mentioned that they would like to use it for other types of analyses that require speaker segmentation. Our tool is based around the LIUM \cite{meignier2010lium} toolkit and just like the previous one, it is completely language independent. Other toolkits were also tested during the project (for example SHoUT\cite{huijbregts2006shout}) but LIUM seemed to perform best on real-world data.

\subsection{Keyword spotting}

Often times, an accurate transcription of audio material is not necessary because we are only interested in individual words appearing in the text. Keyword spotting (KWS) is a process that takes an audio file with a list of keywords and generates a list of their occurrences with in the audio file.

Our system is based around the Kaldi toolkit, but it is also expanded to support an open vocabulary scenario. Given the limited language model vocabulary size, it would be impossible to predict all the words that people may be looking for. Therefore, our system uses a combination of words and syllables, so when a word out of vocabulary needs to be found, its syllable representation is used instead. This makes the tool sometimes more useful than full speech transcription, because it can deal with words that are out-of-vocabulary (OOV), but is more prone to errors when given very short keywords. If the word is phonetically a part of another word it may still be recognized as a separate word.

A small corpus was prepared to test this component and the overall precision was very high ($>\sim$95\%) with the recall being reasonably high for known words ($\sim$82\%) and low for words that were OOV ($\sim$20\%). It seems that the syllable model worked well sometimes but still needs improvement to deal with OOVs.

\begin{figure}
    \centering
\texttt{
\.ze 5.91 0.3 7228.28\\
\.ze 20.21 0.35 5301.86\\
\.ze 20.21 0.13 5266.03\\
\.ze 1.11 0.13 4021.23\\
\.ze 1.23 0.17 4014.55\\
\.ze 0.79 0.12 3494.49\\
\.ze 28.29 0.17 1822.69\\
\.ze 16.6 0.08 0\\
listopada 7.43 0.58 3877.51\\
listopada 29.26 0.5 2541.87\\
polityki 11.27 0.63 7678.28\\
}
    \caption{The output of the keyword spotting tool, searching for the words ``\.ze'', ``listopada'' and ``polityki'' in an example recording. Each line contains one occurrence of a keyword with the following fields: word, start time, duration, keyword likelihood. Note that the short word ``\.ze'' often occurrs as a part of longer words and is thus erroneously detected multiple times.}
\end{figure}

\subsection{Automatic speech transcription}

This tool uses an Automatic Speech Recognition (ASR) system (based on the Kaldi toolkit) to generate a probable orthographic transliteration of audio recording of Polish speech. Initially, this tool was not planned for inclusion in the project but due to overwhelming interest, it was added in the second part of the project. The current system uses our Euronews model for recognizing broadcast news \cite{marasek2014spoken} and in order for it to be useful for other types of recordings, it has to be adapted to the proper domain. More details on the developed architecture is given in section \ref{sec:baseline} below.

\section{Speech Corpus}

In order to produce most of the tools mentioned in the previous section, a large set of good quality recordings is required. This is usually expensive to produce and even if such data is available for purchase from third-parties, it is usually very expensive and unobtainable by most researchers. Prior to our work, there was no free, high quality, large-vocabulary audio corpus of Polish speech. Our goal was to create such a corpus and release it on an open license, both for commercial and non-commercial use.

The corpus was recorded in a studio environment using two microphones: a high-quality studio microphone and a typical consumer audio headset. The corpus consists of 317 speakers recorded in 554 sessions, where each session consists of 20 read sentences and 10 phonetically rich words. The size of the audio portion of the corpus amounts to around 56 hours, with transcriptions containing 356674 words from a vocabulary of size 46361. In addition to the studio corpus, a smaller corpus of telephony quality was also recorded. It contains 114 sessions, amounting to around 13 hours of recorded speech.

Both the studio and telephone quality corpora were released in two forms. The first one is the EMU database \cite{cassidy2001multi}, which allows for easy lookup of data and even some statistics thanks to the integration with the R platform. Unfortunately, the current version of the system relies on downloading the rather sizable corpus locally onto the computer. A new release of the corpus is planned using the more modern EMU Speech Database Management System (EMU-SDMS) that works in the web browser. This will make the corpus much more convenient, since it won't require downloading of all the data. 

\subsection{Baseline speech recognition system}
\label{sec:baseline}

Given that the main purpose of preparing the corpus was the development of speech tools, it seemed fit to deliver the corpus in a form that would make it easy to make such a tool. Since most of the tools mentioned in the previous chapter rely on the Kaldi speech recognition system, a baseline setup for developing such a tool was constructed. This setup was designed to replicate the approach used in the official project for other languages. The main idea is to have all the tools necessary in one folder, with one main script performing the training of the full system, from start to finish. If the user doesn’t wish to modify anything, they can simply run the one script and wait a few hours to get a fully-trained, fully-working large-vocabulary continuous speech recognition system. Users are, however, encouraged to read the comments in the script and figure what is actually being performed.

The corpus was split into a training and test portion, roughly 90\% and 10\% respectively: 56 random sessions were chosen to be in the test set, and the remaining 499 session were stored in the test set. A trigram statistical language model was trained using a large collection of texts in Polish collected from various online sources and interpolated on the transcriptions from the training portion of the corpus only. Since the source material for this portion of the setup cannot be distributed freely, the trained language models are provided as a download.

The acoustic models were trained using the standard GMM training procedure: first the monophone (mono) models were trained, followed by triphone (tri2a). The standard feature front-end is then replaced by an LDA multi-splice set (tri2b), followed by adapting the models to speakers (tri3b). Next, the silence models are are retrained and the phonetic dictionary is rescored (tri3b-sp). Then an experiment is performed with increasing the number of mixtures (tri3b-20k) and with using MMI training (tri3b-mmi). This last one is also repeated using a larger beam, followed by rescoring using a much larger language model, to give the final result of 7.37\% word error rate (WER). If we use the lattice from the wide beam stage and instead of rescoring look for the best sequence of words (in other words, if we had the ideal language model), we can get a score as low as 3.23\% WER. 

In addition to the standard GMM acoustic models, two artificial neural network (ANN) based systems were also tested. The time-delay neural network (TDNN) system achieves a score significantly better than GMMs and the LSTM is even slightly better than that. The LSTM (being a recurrent ANN model) is however much slower to train and the marginal improvement in WER may not be worth it for most people.

\begin{table}
\centering
\begin{tabular}{ll}
\hline
WER \% & experiment               \\
\hline
\hline
30.06 & mono               \\
17.56 & tri1               \\
16.75 & tri2a              \\
15.75 & tri2b              \\
13.50  & tri3b              \\
13.10  & tri3b-sp           \\
12.88 & tri3b-20k          \\
12.41 & tri3b-mmi          \\
11.64 & +wide beam     \\
\textbf{7.37}  & +large LM rescoring \\
3.23 & oracle of wide beam \\
\hline
\end{tabular}
\caption{GMM acoustic model results.}
\label{gmm-res}
\end{table}

\begin{table}
\centering
\begin{tabular}{ll}
\hline
WER \% & experiment               \\
\hline
\hline
9.25   & TDNN                 \\
\textbf{5.91}   & +large LM rescoring \\
2.83 & oracle \\
\hline
8.91   & LSTM                 \\
\textbf{5.78}   & +large LM rescoring \\
2.61 & oracle  \\
\hline
\end{tabular}
\caption{ANN acoustic model results.}
\label{ann-res}
\end{table}

\section{Applications}

A couple of projects have already utilized our tools and resources for their own uses. Our speech alignment tool was used by a consortium partner in order to further annotate the corpora on their Spokes platform \cite{pezik2015spokes}. The studio speech corpus was used in a paper by a Czech research team \cite{nouza2015polish}. We have also managed to cooperate with a team from the Institute of Applied Linguistics at the Warsaw University on their project titled ``Respeaking - the process, competences and quality'' (project code NCN - OPUS6 -2013/11/B/HS2/02762). Finally, one of the most interested groups were researchers of sociology interested in automatic transliteration of sociological interviews. We managed to receive several hours of recordings by a group of researchers from the The Cardinal Wyszy\'{n}ski University in Warsaw. Some preliminary results show promise but more work is needed to achieve success.

We intend to open several new areas of applications in the future. The new project will concentrate mostly around these three domains: parliamentary speeches, historical early and mid-20th century news segments and improved systems for the transliteration of sociological interviews.

\section{Future plans}

With the project being prolonged for the two more years, several improvements are planned. The main focus will be on creating working speech recognition solutions for the aforementioned domains. To achieve this, certain tools, like the G2P conversion including text normalization and possibly other modules, like speaker diarization and VAD, will have to be improved. The biggest improvements, however, will lie in the speech recognition engine, itself. Many experiments are planned, including various adaptation techniques, Deep Neural Network for acoustic modeling \cite{vu2014multilingual}, Recurrent Neural Networks for language modeling \cite{mikolov2013distributed}.

No new corpora will be recorded, although lots of data will have to be collected, in order to adapt the tools to their respective domains. It is unclear whether all of the data will be released for other researchers, due to legal concerns. Our primary intention will be to improve the services available on our website and to provide the trained models and tools for free, for others to use as they deem necessary.

\section{Acknowledgments}

The activities described in this paper were funded by the Clarin-PL project. This project was partially supported by the infrastructure bought within the project ``Heterogenous Computation Cloud'' funded by the Regional Operational Programme of Mazovia Voivodeship.

\bibliography{bibliography}

\end{document}